\PassOptionsToPackage{table}{xcolor}
\documentclass[10pt,journal,compsoc]{IEEEtran}
\usepackage[
left=0.45in,
right=0.45in,
top=0.48in,
bottom=0.48in,
showframe=false 
]{geometry}
\usepackage[utf8]{inputenc}
\usepackage[T1]{fontenc}        
\usepackage{newtxtext,newtxmath} 
\usepackage{setspace}
\usepackage{xcolor} 
\usepackage{graphicx}
\graphicspath{{Fig/}}
\DeclareGraphicsExtensions{.pdf,.png,.jpg}

\usepackage[caption=false,font=footnotesize,labelfont=sf,textfont=sf]{subfig}

\usepackage{capt-of}

\usepackage{booktabs}
\usepackage{multirow}
\usepackage{array}
\usepackage{makecell}
\usepackage{colortbl} 

\definecolor{mygray}{gray}{.55}
\definecolor{bestbg}{RGB}{255, 235, 235}
\definecolor{secondbg}{RGB}{240, 248, 255}
\definecolor{rowgray}{gray}{0.95}

\newcommand{\best}[1]{\textbf{#1}}
\newcommand{\second}[1]{\underline{#1}}

\usepackage{bm}
\usepackage{mdwmath}
\usepackage{eqparbox}
\usepackage{marvosym} 
     
\usepackage{bbding}
\usepackage{pifont} 
\DeclareMathOperator{\Normalize}{normalize}

\usepackage{tikz}
\usepackage{pgfplots}
\pgfplotsset{compat=1.12}

\usepackage{url}
\usepackage{algorithm}
\usepackage{algpseudocode}
\usepackage{footnote}
\usepackage{pdflscape}
\usepackage{afterpage}

\usepackage[square,sort,comma,numbers,sort&compress]{natbib}
\usepackage[pagebackref=false,breaklinks=false,linkcolor=red,anchorcolor=black, citecolor=black,colorlinks,bookmarks=true]{hyperref}

\newcommand{\cmark}{\ding{51}} 
\newcommand{\xmark}{\textcolor{gray}{\ding{55}}} 

\begin{document}
	\title{
		Towards UAV Detection in the Real World: A New Multispectral Dataset UAVNet-MS and a New Method}	
	\author{Yihang~Luo, Jun~Chen, Chao~Xiao, Yingqian~Wang, Zhaoxu~Li, Qiang~Ling, Xu~He, Nuo~Chen, Gaowei~Guo, Hongge~Li, Miao~Li, Longguang~Wang, Yulan~Guo, Li~Liu, Wei~An, and Zhijie~Chen
		\IEEEcompsocitemizethanks{
		\IEEEcompsocthanksitem This work was supported by the National Natural Science Foundation of China under Grants 62401591, 62501609, and 62401590. 
	Y.~Luo (luoyihang@nudt.edu.cn), J.~Chen, C.~Xiao (xiaochao12@nudt.edu.cn), Y.~Wang, Z.~Li, Q.~Ling (lingqiang16@nudt.edu.cn), X.~He, N.~Chen, G.~Guo, H.~Li, M.~Li, L.~Liu (liuli\_nudt@nudt.edu.cn), and W.~An are with the College of Electronic Science and Technology, National University of Defense Technology, Changsha 410073, China.
	L.~Wang is with the Aviation University of Air Force, China.
	Y.~Guo is with Sun Yat-sen University, China.
    Q.~Ling, C.~Xiao, and L.~Liu are the corresponding authors.
	}
	}

	\markboth{Submitted to IEEE Transactions on Pattern Analysis and Machine Intelligence}%
	{Luo \MakeLowercase{\textit{et al.}}: UAVNet-MS}
	
	\IEEEtitleabstractindextext{
		\begin{center}
			\vspace{-0.6cm} 
			\setcounter{figure}{0}
			\includegraphics[width=0.92\textwidth]{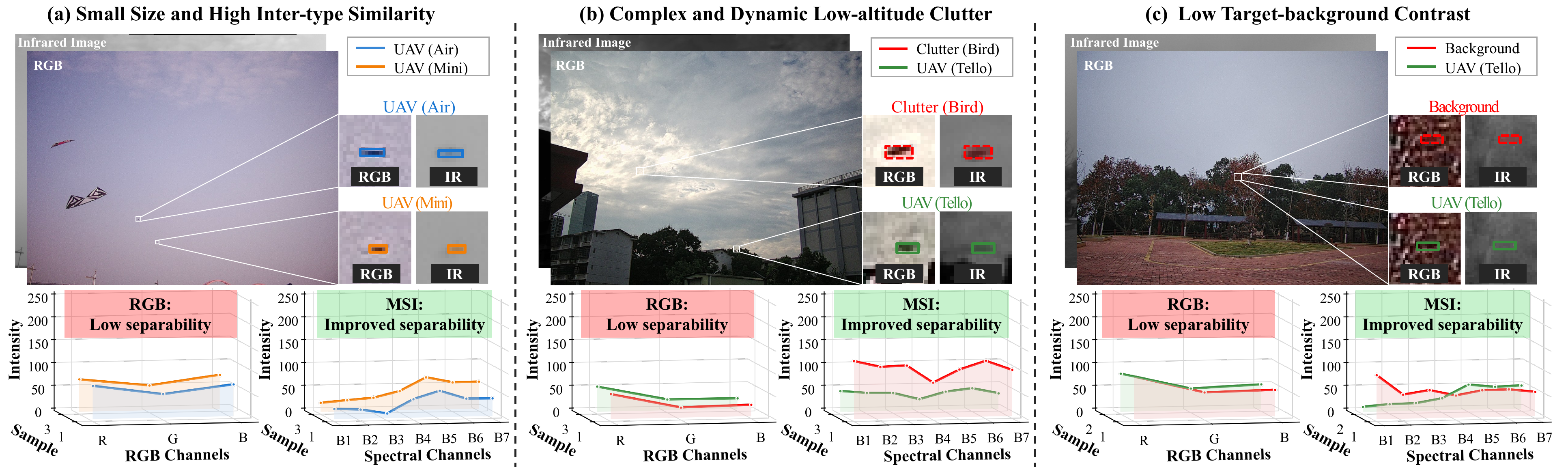}
		\captionof{figure}{Three key challenges in fine-grained small-UAV detection and the motivation for spectral cues. 
			Each case compares RGB with an infrared (IR) image example (top) and the corresponding separability across RGB channels and MSI bands (bottom).
			In challenging scenes, spatial cues can be unreliable due to small-scale inter--type similarity among UAVs (a), confusion with dynamic clutter such as birds (b), and low target--background contrast against complex backgrounds (c).
			In contrast, multispectral signatures provide material-dependent cues that improve separability against background/clutter and across material-diverse UAV types.
			Here, $B_i$ denotes the $i$-th MSI band.}				
			\label{motivation}
	 	\vspace{0.2cm} 
		\end{center}		
		\begin{abstract}
		The proliferation of unmanned aerial vehicles (UAVs) within the low-altitude economy has created an urgent demand for precise, fine-grained UAV monitoring systems. 
		Despite recent progress, existing UAV monitoring systems based on RGB imaging or RGB paired with infrared (IR) fusion predominantly rely on spatial cues, whose discriminative ability degrades at small scales, particularly in the presence of high inter--type similarity, target--clutter ambiguity (e.g., birds), and low target--background contrast.
		This limitation motivates the use of complementary information grounded in physical properties. Multispectral imaging (MSI) encodes material-aware spectral signatures, yet MSI-based fine-grained small-UAV detection remains underexplored, largely due to the absence of dedicated datasets, baselines and benchmarks for evaluation.
		To fill this gap, we introduce UAVNet-MS, the first multispectral dataset tailored for fine-grained small-UAV detection.
		UAVNet-MS comprises 15,618 temporally synchronized RGB--MSI data cubes with high spatial resolution ($1440\times1080$), captured by an array-based multispectral imaging system and annotated with bounding boxes and four material-diverse UAV types.
		UAVNet-MS also presents a demanding small-object profile (93.7\% of targets have areas $\le 32^2$ pixels; the average target area is at most $18^2$ pixels, occupying only $\sim$0.02\% of the image area) under pervasively low contrast, to better reflect challenging real-world low-altitude UAV monitoring scenarios.
		Based on the UAVNet-MS dataset, we propose the multispectral fusion detector (MFDNet), a dual-stream baseline tailored to address array-induced inter-band parallax and spatial--spectral fusion for fine-grained UAV detection.
		Through extensive evaluation under three standardized protocols (RGB-only, MSI-only, RGB+MSI) and comprehensive comparisons with 20 detectors, MFDNet achieves a significant gain of +6.2\% AP$_{50}$ over the best RGB-only detector, showing that spectral cues provide complementary material evidence that strengthens discrimination beyond spatial cues alone.
		Overall, our work provides a foundational dataset, a strong baseline, and a benchmark to advance future research in multispectral UAV monitoring.
		\end{abstract}
	}
	\maketitle
	\IEEEdisplaynontitleabstractindextext
	\IEEEpeerreviewmaketitle
	
	\IEEEraisesectionheading{\section{Introduction}\label{sec:introduction}}
\IEEEPARstart{T}{he} low-altitude economy (LAE) is rapidly emerging as a new strategic engine for regional development and industrial upgrading, demonstrating broad application potential and significant economic value~\cite{9573394}. 
As an indispensable component of the LAE, unmanned aerial vehicles (UAVs) are being deployed pervasively across civilian and military domains, enabling applications ranging from precision mapping and logistics to surveillance~\cite{Anti-uav410-6,10004511}.
However, the proliferation of UAVs intensifies pressing challenges in low-altitude airspace management and raises public safety and security concerns~\cite{Robot2025overview,liu2026atrnet}.
This underscores the urgent need for precise and fine-grained UAV monitoring systems capable of operating in complex low-altitude environments, thereby supporting informed response decisions.
Despite this clear importance, achieving reliable UAV monitoring under realistic low-altitude conditions remains challenging, cementing its status as a vital yet underexplored research frontier.

Existing UAV monitoring systems have made considerable progress, focusing on improving UAV localization and false-alarm suppression using RGB imaging or RGB paired with infrared (IR) fusion~\cite{RGBT234-25,VTUAVdalian-20,Anti-UAV-24}. While effective, these systems rely mainly on spatial cues (shape, texture, and intensity patterns), whose discriminative power is challenged in low-altitude fine-grained detection in three aspects:

%
%
(1) Target-related challenges: small size and high inter--type similarity. 
UAVs often appear as small objects (only a few pixels across), where discriminative spatial cues (e.g., shape and texture) become indistinct, making accurate localization fundamentally challenging.
Moreover, this difficulty is compounded by high inter--type similarity among diverse UAV types, as fine-grained differences collapse at such scales, leading to misclassification even when the target is correctly localized (Fig.~\ref{motivation}(a)).

(2) Environment-related challenges: complex and dynamic low-altitude clutter. 
Low-altitude monitoring must contend with diverse and non-stationary backgrounds, where dynamic clutter is prevalent.
As shown in Fig.~\ref{motivation}(b), visually similar moving clutter, such as birds, can closely resemble distant UAVs in apparent shape and intensity patterns under a large field-of-view.
This ambiguity leads to frequent false alarms, making robust false-alarm suppression difficult when relying solely on spatial cues.

(3) Target--environment coupling challenge: low target--background contrast. 
The most critical issue arises from the adverse coupling between feature-deficient small targets and complex backgrounds, which jointly diminishes target saliency and makes spatial cues unreliable.
As shown in Fig.~\ref{motivation}(c), UAVs can visually blend into backgrounds such as foliage or building facades due to similar colors and textures, yielding severely low target--background contrast and posing a fundamental challenge to reliable detection.

Collectively, these challenges underscore that reliance on spatial cues alone is inherently inadequate for robust, fine-grained UAV detection. This fundamental limitation motivates the pursuit of complementary information grounded in physical properties. 
Multispectral imaging (MSI) captures scene reflectance across multiple narrow spectral bands and thus encodes material-dependent spectral signatures, providing discriminative evidence beyond spatial cues~\cite{Material-8,WHU-Hi-H3-16,MUST}.
However, despite this potential, fine-grained small-UAV detection with material-aware spectral cues remains underexplored, primarily due to the lack of dedicated datasets, baselines and benchmarks for evaluation.
\begin{figure}[t]
	\captionsetup {font=scriptsize, labelfont=scriptsize}
	\centering
	\vspace{-0.2cm} 
	\includegraphics[width=0.45\textwidth]{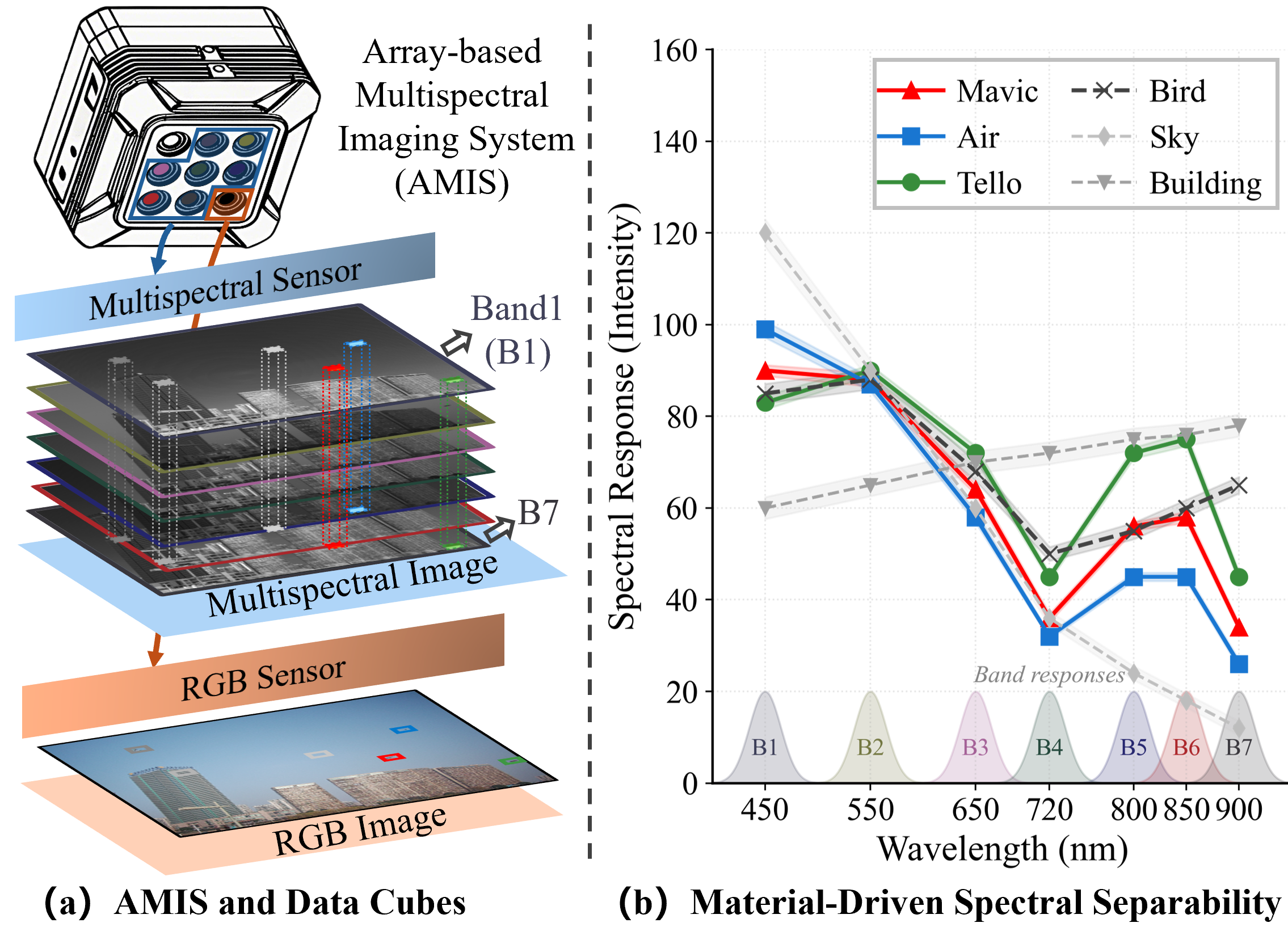}
		\vspace{-0.3cm}
	\caption{AMIS imaging system and spectral separability. 
    (a) Example of a 7-band MSI cube and an RGB image captured by AMIS with UAV annotations. 
(b) Spectral response curves (mean ± 1 standard deviation) of three material-diverse UAV types (solid lines) and three background/clutter classes (dashed lines).}
    \label{collect}
	\vspace{-.6cm}
\end{figure}

To bridge these gaps, we advance fine-grained small-UAV detection through three pivotal contributions: (i) a dedicated multispectral dataset (UAVNet-MS) with systematic analyses, (ii) a principled RGB--MSI baseline detector (MFDNet), and (iii) standardized evaluation protocols and extensive experiments that quantify multispectral benefits and support comparisons.

\begin{figure}[t]
	\captionsetup {font=scriptsize, labelfont=scriptsize}
	\centering
	\vspace{-0.2cm}
	\includegraphics[width=0.45\textwidth]{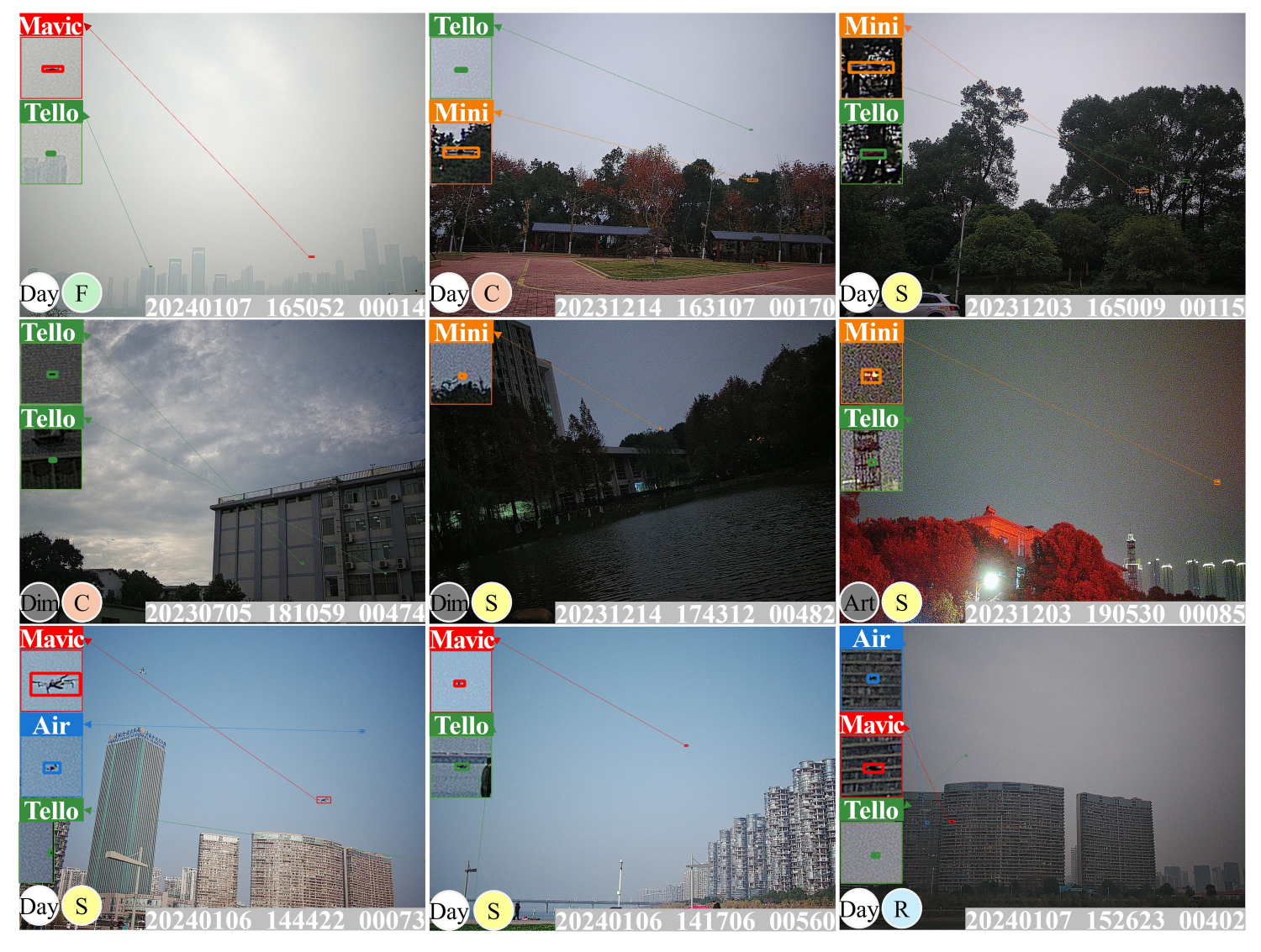}
	\vspace{-0.4cm} 
	\caption{Environmental diversity in UAVNet-MS dataset. 
		Two circular markers encode illumination (\textit{i.e.}, Day: daylight, Dim: dim natural light, and Art: artificial lighting) and weather (\textit{i.e.}, S: sunny, R: rainy, C: cloudy, and F: foggy).
		UAV types are color-coded: \textcolor{red}{Mavic}, \textcolor{green}{Tello}, \textcolor{orange}{Mini}, and \textcolor{blue}{Air}.} \label{Scene}
	\vspace{-0.5cm}
\end{figure}

As a first step toward bridging the multispectral data gap in low-altitude UAV detection, we present UAVNet-MS, a material-aware multispectral dataset for fine-grained small-UAV detection that provides temporally synchronized, high-resolution RGB and MSI imagery and complements prior UAV detection datasets in Table~\ref{tab:related_dataset1}.
UAVNet-MS comprises 15,618 RGB--MSI data cubes (3-channel RGB plus 7-band MSI), using an array-based multispectral imaging system (AMIS, Fig.~\ref{collect}(a)) at up to 25~fps. All cubes are temporally synchronized, share a $1440\times1080$ spatial resolution, and are annotated with four UAV types of distinct material compositions.
Compared with existing resources, our custom AMIS gives UAVNet-MS temporally synchronized, high-resolution cubes tailored to small-UAV detection, a combination rare in UAV datasets and often sacrificed in MSI collections.
With such sensing fidelity, the four material-diverse UAV types yield divergent spectral signatures (Fig.~\ref{collect}(b)), enabling evaluation of material-aware, fine-grained separability beyond spatial cues.
Moreover, UAVNet-MS spans 4 weather conditions, 15 scene types, and varying illumination (Fig.~\ref{Scene}), capturing core low-altitude challenges.
This yields a demanding detection profile: 93.7\% of targets are small (area $\le 32^2$ pixels), averaging at most $18^2$ pixels (only $\sim$0.02\% of image area) under low contrast.
Overall, UAVNet-MS required a year of meticulous system design, data acquisition, processing, and annotation, demonstrating substantial investment to ensure data quality and relevance.

Based on the UAVNet-MS dataset, we propose the multispectral fusion detector (MFDNet), a dual-stream baseline tailored to address array-induced inter-band parallax and spatial--spectral fusion for fine-grained UAV detection.
Specifically, MFDNet introduces a shared array-aware positional encoding module (ArrayCode) that injects camera-array geometry into both modalities, enabling the network to learn and compensate for the inter-band misalignment inherent in array-based acquisition.
To preserve complementary cues, MFDNet adopts modality-specific backbones to separately encode RGB spatial textures and MSI inter-band spectral correlations.
Moreover, MFDNet employs a scale-aware semantic-decoupled fusion strategy that selectively injects fine-scale spectral cues to enhance small-target contrast, while preserving a stable, high-level RGB semantic pathway for robust false-alarm suppression.
By unifying these components into an end-to-end trainable framework, MFDNet provides a strong and practical baseline for UAVNet-MS, thereby enabling reliable fine-grained UAV detection with array-based multispectral sensing.

To enable fair evaluation, we define three protocols (RGB-only, MSI-only, and RGB+MSI) and conduct extensive experiments, including benchmarking against 20 representative detectors, to validate the baseline and quantify the benefits of multispectral cues for fine-grained small UAV detection.
Our contributions can be summarized as follows:

(1) We establish a dedicated benchmark setting for fine-grained small-UAV detection with multispectral cues. To support it, we introduce UAVNet-MS, the first material-aware multispectral dataset tailored for this setting, providing temporally synchronized and resolution-consistent RGB--MSI data across four material-diverse UAV types under realistic low-contrast conditions and enabling analysis of spectral cues for small UAVs.

(2) We propose MFDNet, a dual-stream detector that compensates for array-induced inter-band parallax via ArrayCode and performs semantic-decoupled fusion to effectively integrate material cues, establishing a solid baseline for RGB--MSI small-object detection.

(3) Through extensive experiments and ablations on UAVNet-MS under standardized protocols, we quantify the empirical benefits of multispectral information and provide baselines and insights to facilitate future MSI-based UAV monitoring research.
\begin{table*}[t]
	\captionsetup{font=scriptsize, labelfont=scriptsize}
	\scriptsize
	\centering
	\vspace{-0.2cm} 
	\caption{Comparison of representative related datasets.
		"Modality" and "Task" represent the dataset sensor and the task type (D: detection, T: tracking).
		"\#UAV Type", "Avg. Size" and "Size. R." denote the number of annotated UAV types, the average target size in pixels, and the target area ratio, respectively.
		"\#Samples (K)" reports the sample count in thousands: RGB/IR (images), RGB-IR (pairs), and MSI (cubes).
		"Resolution" and "Max FPS" denote the maximum spatial resolution and maximum frame rate, respectively.
		For MSI, "Spec. R." indicates the spectral range (nm).
		"Scenario" denotes the primary scenario, and "Scn. D." indicates the scene diversity.
		"MR Con.", "Align", and "LL" represent multi-modal resolution consistency, cross-modal alignment, and low-light support (Y: yes, N: no).
	}\label{tab:related_dataset1}
	\vspace{-.3cm}
	\setlength{\tabcolsep}{0.8mm}
	\renewcommand\arraystretch{.98}
	\begin{tabular}{cccccccccccccccc}
		\hline
		\textbf{Benchmark} & \textbf{Modality} & \textbf{Task} & \textbf{\#UAV Type} & \textbf{Avg. Size} & \textbf{Size. R.} & \textbf{\#Samples (K)} & \textbf{Resolution} & \textbf{Max FPS} & \textbf{Spec. R.} & \textbf{Scenario} & \textbf{Scn. D.} & \textbf{MR Con.} & \textbf{Align} & \textbf{LL} & \textbf{Year} \\\hline
		UAV123 \cite{UAV123} & RGB & T & - & 84$\times$42 & 0.46 & 113 & 1280$\times$720 & 30 & - & Outdoor & Y & - & - & N & 2016 \\
		DTB70 \cite{DTB70} & RGB & T &  - & 76$\times$38 & 0.40 & 16 & 1280$\times$720 & 30 & - & Outdoor & Y & - & - & N & 2017 \\
		VisDrone \cite{VisDrone} & RGB & D & - & 45$\times$25 & 0.09 & 261 & 1920$\times$1080 & 25 & - & Urban & Y & - & - & N & 2021 \\
		UAVDT \cite{UAVDT} & RGB & D\&T & - & 58$\times$32 & 0.16 & 80 & 1080$\times$540 & 30 & - & Urban & Y & - & - & N & 2018 \\
		StanfordDrone \cite{StanfordDrone} & RGB & D\&T & - & 62$\times$35 & 0.18 & 60 & 1920$\times$1080 & 30 & - & Campus & Y & - & - & N & 2016 \\
		MAV-VID \cite{MAV-VID-59} & RGB & D & 1 & 136$\times$77 & 0.54 & 40 & 1920$\times$1080 & 30 & - & Central & N & - & - & N & 2020 \\
		TJU-DHD \cite{TJU-DHD} & RGB & D & - & 49$\times$28 & 0.12 & 60 & 1624$\times$1200 & - & - & Traffic & Y & - & - & N & 2021 \\
		AntiUAV410 \cite{Anti-uav410-6} & IR & T & 1 & 40$\times$40 & 0.80 & 18.6 & 640$\times$512 & 30 & - & Outdoor & Y & - & - & N & 2024 \\
		1st-AntiUAV \cite{Anti-UAV-24} & RGB-IR & D\&T & 1 & 45$\times$45 & 0.62 & 586 & 640$\times$512/1920$\times$1080 & 20 & - & Outdoor & N & N & N & N & 2020 \\
		VTUAV \cite{VTUAVdalian-20} &  RGB-IR & T &  - & 52$\times$48 & 0.76 & 1700 & 1920$\times$1080 & 30 & - & Aerial & Y & Y & Y & Y & 2022 \\
		KAIST \cite{KAIST} &  RGB-IR & D & - & 68$\times$42 & 0.47 & 95 & 640$\times$480/1920$\times$1080 & 20 & - & Urban & N & N & Y & Y & 2015 \\
		FLIRv2 \cite{FLIR-28} &  RGB-IR & D & - & 72$\times$45 & 0.51 & 26 & 640$\times$512/1920$\times$1080 & 24 & - & Driving & N & N & N & Y & 2020 \\
		LLVIP \cite{Llvip-27} &  RGB-IR & D & - & 65$\times$38 & 0.34 & 31 & 1080$\times$720/1920$\times$1080 & - & - & Night & N & N & N & Y & 2021 \\
		LasHeR \cite{LasHeR-26} &  RGB-IR & T &  - & 74$\times$46 & 0.53 & 6700 & 630$\times$480/1920$\times$1080 & 20 & - & Mixed & Y & N & N & Y & 2021 \\
		RGBT234 \cite{RGBT234-25} &  RGB-IR & T &  - & 69$\times$43 & 0.48 & 234 & 630$\times$460/1920$\times$1080 & 30 & - & Various & Y & N & Y & Y & 2019 \\
		RGBT-Tiny \cite{Tiny24-XY} &  RGB-IR & D & 1 & 51$\times$31 & 0.31 & 93 & 640$\times$512/1920$\times$1080 & 15 & - & Various & Y & - & - & - & 2024 \\
		VEDAI \cite{VEDAI} &  RGB-IR & D & - & 85$\times$52 & 0.72 & 0.6 & 512$\times$512/1024$\times$1024 & - & - & Aerial & N & N & N & N & 2016 \\
		RooftopHSI \cite{RooftopHS-14} & HSI & D & 1 & - & - & 1.6 & 1600$\times$192 & 1.2 & 390-900 & Driving & N & - & - & N & 2022 \\
		HOT \cite{Material-8} & MSI & T & - & - & - & 21 & 512$\times$256 & 25 & 470-620 & Indoor & Y & - & - & Y & 2020 \\
		H3 \cite{WHU-Hi-H3-16} & MSI & T & - & - & - & 23 & 512$\times$256 & 25 & 470-620 & Life & Y & N & N & Y & 2022 \\
		MUST \cite{MUST} & MSI & T & - & - & - & 43 & 1200$\times$900 & 5 & 450-900 & Aerial & Y & Y & Y & Y & 2024 \\
		\hline
		\rowcolor{gray!8}
		\textbf{UAVNet-MS} & \textbf{MSI} & \textbf{D\&T} & \textbf{4} & \textbf{18$\times$18} & \textbf{0.02} & \textbf{15.6} & \textbf{1440$\times$1080} & \textbf{25} & \textbf{450-900} & \textbf{Urban} & \textbf{Y} & \textbf{Y} & \textbf{Y} & \textbf{Y} & \textbf{2025} \\\hline
	\end{tabular}
	\vspace{-.4cm}
\end{table*}
\section{Related Work}
\label{relatedwork}
\subsection{Datasets: From Spatial to Spectral UAV Monitoring}
\label{sec:related_data}
As summarized in Table~\ref{tab:related_dataset1}, the evolution of UAV monitoring related datasets reflects a progression from purely spatial sensing to multi-modal fusion. 

\noindent\textbf{Spatial-based Benchmarks (RGB/Infrared).}
Existing datasets are mainly divided into UAV-centric datasets~\cite{UAV123,DTB70,VisDrone,UAVDT,11112674,StanfordDrone,MAV-VID-59,TJU-DHD} that image ground scenes, and UAV monitoring datasets~\cite{Anti-uav410-6,Anti-UAV-24} targeting UAVs themselves.
However, the former focuses on general ground objects under viewpoints that differ from low-altitude UAV monitoring, while the latter often lacks fine-grained type labels for UAVs. This limitation leaves a clear gap for datasets supporting small, fine-grained UAV detection in realistic scenes.

\noindent\textbf{Environment-Adaptive Benchmarks (RGB-Infrared).}
To mitigate environmental interference, RGB and Infrared (RGB-IR) bi-modal datasets~\cite{Llvip-27,FLIR-28,11342305,LasHeR-26,RGBT234-25,VTUAVdalian-20,KAIST,VEDAI,Tiny24-XY} introduce thermal sensors to enhance robustness.
While effective under varying illumination, RGB and thermal signals are often insufficient to separate UAVs from background clutter with similar appearance or temperature.

\noindent\textbf{Material-Aware Benchmarks (MSI/HSI).}
Spectral datasets offer the potential for material identification but face hardware and task-specific limitations.
Pushbroom systems~\cite{RooftopHS-14} suffer from motion-induced distortions, while snapshot approaches~\cite{Material-8,WHU-Hi-H3-16} sacrifice spatial resolution and SNR, both of which are critical for identifying small objects.
Although the MUST dataset~\cite{MUST} takes a step toward spectral perception, it is designed for single-object tracking and lacks multi-class annotations. Consequently, the role of MSI in fine-grained, small-scale UAV detection remains unexplored mainly due to the absence of a dedicated benchmark. This gap motivates our introduction of the UAVNet-MS dataset.
 \vspace{-0.2cm} 
\subsection{Methodologies: From RGB Detection to Spectral Fusion}
\noindent\textbf{Appearance-based Detection in RGB.}
Mainstream RGB detectors~\cite{FasterR-CNN,FCOS,chen2025motion,CenterNet,DeformableDETR,DINO} and slicing schemes~\cite{SAHI} have advanced the field.
However, because these methods fundamentally rely on spatial cues, their performance saturates once objects shrink to only a few pixels and lose distinctive shape cues.

\noindent\textbf{Multi-modal Fusion Strategies.}
While  RGB-IR fusion methods~\cite{LasHeR-26,RGBT234-25,VTUAVdalian-20,11098674} improve detection robustness under illumination variations, their discriminative capability still primarily relies on spatial cues (e.g., contrast, texture, and shape).
Recent spectral tracking advances~\cite{RooftopHS-14,Material-8,MUST} validate the robustness of material cues against occlusion and appearance degradation.
Despite this success in tracking, detection methodologies tailored explicitly for small objects in the spectral domain remain largely unexplored. This paper bridges this gap by proposing an RGB--MSI dual-stream detection baseline.
\begin{figure*}[t]
	\captionsetup {font=scriptsize, labelfont=scriptsize}
	\centering
	\vspace{-0.2cm} 
	\includegraphics[width=0.9\textwidth]{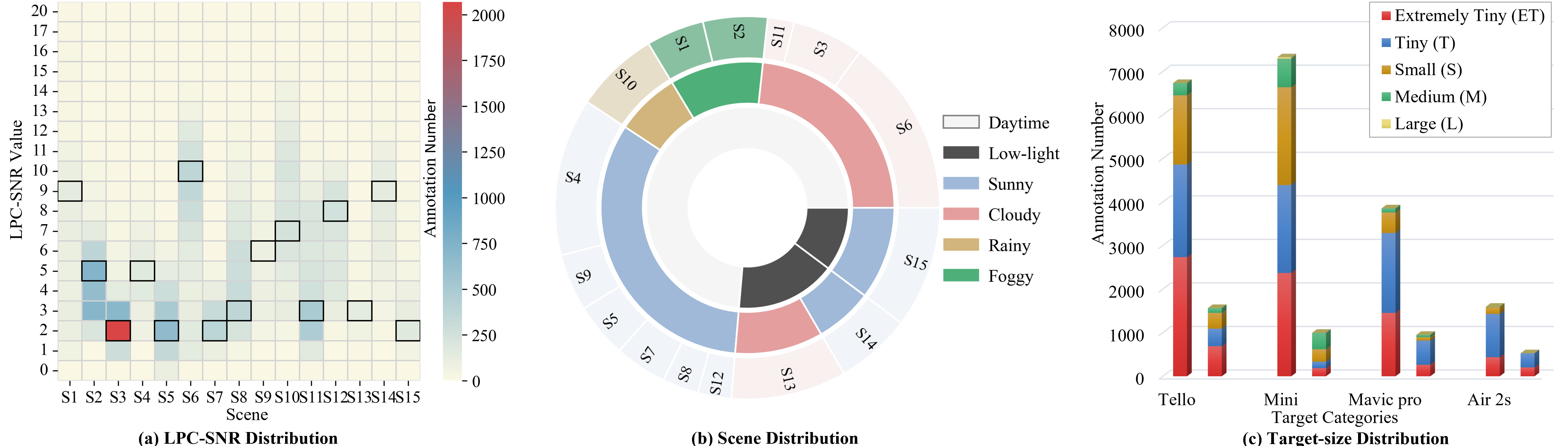}
			\vspace{-0.2cm}
	\caption{Statistics of the UAVNet-MS dataset. 
(a) Local peak-contrast SNR (LPC-SNR) distribution across different scene attributes. 
Black boxes mark the dominant LPC-SNR range of targets for each scene condition.
(b) Scene distribution across illumination conditions, weather patterns, and background types.
(c) Target size distribution bins (ET/T/S/M/L), where 93.7\% of instances fall into ET/T/S. 
    }\label{fig:anly}
	\vspace{-.55cm}
\end{figure*}
 \vspace{-0.4cm} 
\section{The UAVNet-MS Dataset}
\subsection{Data Collection and Annotation}
\textbf{Data Capture.}
The UAVNet-MS dataset is acquired using an AMIS, which synchronously captures high-resolution RGB frames and seven narrow spectral bands spanning 450--900~\,nm.
Both modalities share a spatial resolution of $1440 \times 1080$ pixels. 
To validate fine-grained separability, we selected four UAV types with distinct material compositions (e.g., carbon fiber vs. plastic) and appearance (e.g., color). Detailed specifications of AMIS and UAVs are provided in the Appendix Sec.~S1.1 and Sec.~S1.2.

\noindent\textbf{Data Preprocessing.}
We adopt a three-stage preprocessing pipeline to ensure data quality and cross-modal consistency.
(i) Geometric registration: Each spectral band is registered to the RGB reference camera using homography-based warps.
(ii) Radiometric quality control: Images with severe signal anomalies or sensor artifacts (e.g., incomplete imaging or pronounced striping noise) are filtered out.
(iii) Temporal synchronization: We verify temporal consistency across cameras using target positions and discard cubes exhibiting noticeable RGB--MSI desynchronization.
More detailed preprocessing steps, together with the calibration procedure and quantitative evaluations of alignment accuracy, are provided in Appendix Sec.~S1.3.

\noindent\textbf{Ground-truth Annotations.} 
We use expert manual labeling on RGB frames to produce ground-truth annotations, with automated transfer to all spectral bands using the geometric transforms estimated during preprocessing.
All 15,618 cubes (containing 26,890 instances) are manually inspected as a quality control step to ensure bounding box precision.
The dataset is partitioned into a 70\%/30\% train/test split, stratified by scene and class to preserve their distributions.
 \vspace{-0.3cm} 
\subsection{Dataset Properties and Statistics}
\label{datasetsatis}
We analyze UAVNet-MS to highlight key challenges in small-scale fine-grained UAV detection. Statistical characteristics observed from UAVNet-MS are summarized as follows:

\noindent\textbf{Prevalence of Low Local Contrast.}
We quantify target visibility using the Local Peak-Contrast SNR (LPC-SNR)~\cite{li2023direction}:
\begin{equation}
\text{LPC-SNR} = \frac{\left| \max_{p \in \Omega} I(p) - \mu_N \right|}{\sigma_N},
\end{equation}
where $\Omega$ denotes the target region, and $N$ represents the immediate background annulus (defined by a 10-pixel dilation of $\Omega$). $\mu_N$ and $\sigma_N$ are the background mean and standard deviation.
A lower LPC-SNR indicates that the target peak is close to the background distribution, reflecting weak local contrast.
The predominance of LPC-SNR values below 10 (Fig.~\ref{fig:anly}(a)) reveals the severe challenge of low target-to-background contrast inherent in the UAVNet-MS dataset.
Such regimes, where spatial cues may become ambiguous (Fig.~\ref{motivation}(a)), provide a suitable testbed to investigate the efficacy of spectral cues.

\noindent\textbf{Diverse Scenes and Dynamic Clutter.}
UAVNet-MS spans diverse environmental conditions (Fig.~\ref{fig:anly}(b)), including various weather patterns and lighting scenarios, with 26.34\% of the data collected under low-light conditions. Such cluttered scenes exacerbate target--clutter ambiguity and false alarms, consistent with Fig.~\ref{motivation}(b).

\noindent\textbf{Dominance of Small Targets.}
Following the protocols in \cite{Tiny24-XY}, we divide targets into five size tiers: 
Extremely Tiny (ET, $[1^2, 8^2)$), 
Tiny (T, $[8^2, 16^2)$), 
Small (S, $[16^2, 32^2)$), 
Medium (M, $[32^2, 96^2)$), 
and Large (L, $\ge 96^2$). 
As shown in Fig.~\ref{fig:anly}(c), targets are dominated by the ET/T/S size tiers, which together account for 93.7\% of all instances.
With an average target size of only $18 \times 18$ pixels ($\approx 0.02\%$ of the image area), the dataset poses a significant challenge due to inadequate spatial resolution.
 \vspace{-0.2cm} 
\subsection{Material-Driven Spectral Separability}
\begin{figure}[th]
	\captionsetup {font=scriptsize, labelfont=scriptsize}
	\centering
	\vspace{-.4cm}
	\includegraphics[width=0.47\textwidth]{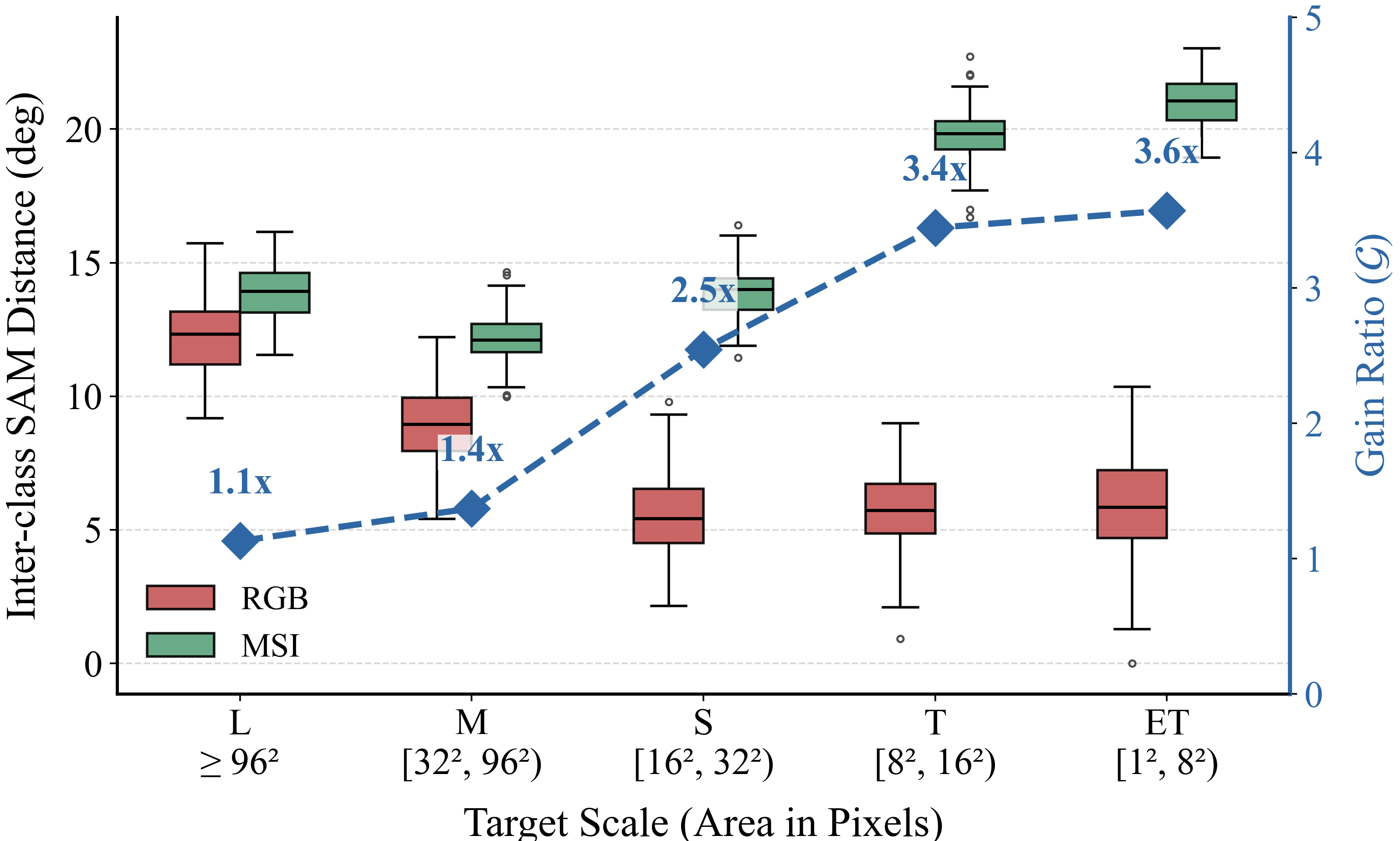}
    \vspace{-.4cm}
	\caption{Inter-type spectral separability across UAV scales. Boxplots show the distribution of pairwise spectral angle mapper (SAM, a measure of spectral shape similarity \cite{li2025specdetr}) between class centers for RGB (blue) and MSI (red) at five scales.
The green dashed line (right axis) indicates the gain ratio 
$\mathcal{G} = \mathrm{SAM}_{\mathrm{MSI}} / \mathrm{SAM}_{\mathrm{RGB}}$.}
\label{fig:sam}
	\vspace{-.4cm}
\end{figure}
We articulate the necessity of multispectral information through an analysis of detection capability (localization) and an empirical validation of material discriminability (classification).

\noindent\textbf{Gain in Localization (Contrast Aggregation).}
Multispectral imaging aggregates contrast-to-noise ratios across bands, 
yielding detection gain $\mathcal{G} \geq 1$  over an arbitrary single band 
(see the Appendix Sec.~S2.2 for derivation).
This complementarity suppresses false alarms from background and 
clutter (Figs.~\ref{motivation}(a), ~\ref{motivation}(b)).

\noindent\textbf{Gain in Classification (Spectral Fingerprints).}
The stability of spectral signatures acts as a unique fingerprint for fine-grained classification. 
Unlike spatial features, which are unstable at small scales, material-dependent reflectance provides more consistent inter-class separability.
As shown in Fig.~\ref{fig:sam}, 
in the ET regime, where targets occupy fewer than $8 \times 8$ pixels, RGB separability degrades sharply due to the loss of spatial shape and texture information.
Conversely, MSI maintains high inter-class distances, supporting that spectral cues provide discriminative power precisely where spatial features fail.
 \vspace{-0.3cm} 
\section{The MFDNet Baseline Detector}
\begin{figure*}[t]
\captionsetup {font=scriptsize, labelfont=scriptsize}
    \centering
    \vspace{-0.2cm} 
    \includegraphics[width=0.9\textwidth]{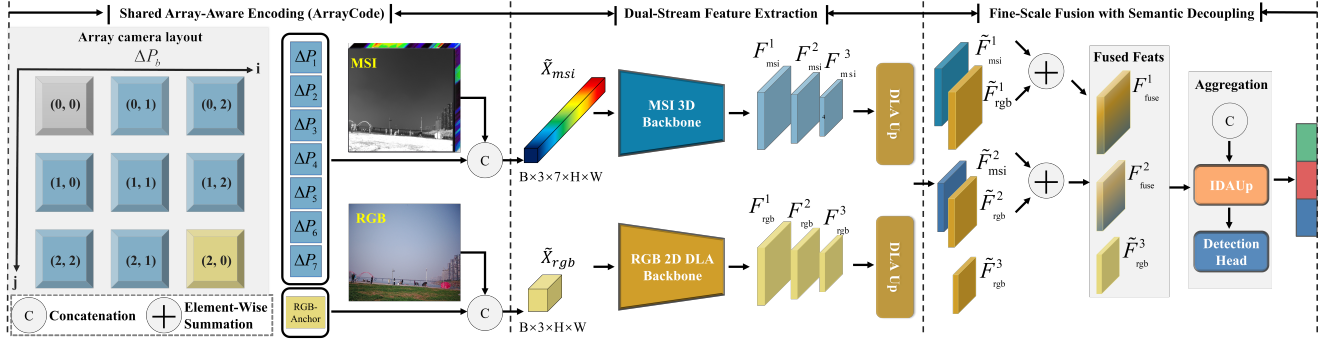}
    		\vspace{-0.5cm}
\caption{Overview of MFDNet. ArrayCode, dual-stream feature extraction, and fine-scale fusion with semantic decoupling are the three key components. 
	First, the ArrayCode module injects geometric priors into the RGB and MSI inputs to resolve array-induced parallax. 
	Then, dual-stream backbones are utilized to independently extract spatial textures and volumetric spectral correlations, preventing feature suppression. 
	Finally, the decoupled features are integrated via fine-scale fusion to predict classification heatmaps, center offsets, and object sizes for robust small object detection.}
    \label{fig:mfdnet}
	\vspace{-.4cm}
\end{figure*}
\subsection{Overall Architecture}
Multispectral Fusion Detector (MFDNet) is a dual-stream detector tailored to array-based sensing and small object detection. In array-based MSI, band-wise parallax and modality imbalance pose two coupled challenges. First, small UAVs are highly sensitive to residual misalignment, and strong RGB textures tend to suppress subtle spectral cues. Moreover, deep fusion can destabilize high-level semantics that are critical for false-alarm suppression.
To cope with these issues, we design MFDNet around three domain-specific principles, as summarized in Fig.~\ref{fig:mfdnet}. These principles are:
(i) \textit{Array-aware geometric prior}: Small targets are highly sensitive to residual parallax across the AMIS sub-cameras. Instead of heavy feature-level warping, we encode the fixed $3\times3$ array layout into an array-aware positional encoding (ArrayCode) module for both modalities.
(ii) \textit{Modality-specific representation learning}: To prevent high-frequency RGB textures from suppressing spectral cues, we use separate backbones that explicitly decouple RGB and MSI features.
(iii) \textit{Scale-aware integration}: We restrict cross-modal fusion to fine-scale layers, ensuring that spectral cues enhance localization without corrupting the high-level semantic abstraction required for false-alarm suppression.

In summary, MFDNet provides an array-aware and semantically stable fusion baseline, bridging the gap between physical MSI sensing and fine-grained small-UAV detection. Ablation studies in Sec.~\ref{Ablation} further validate that each design choice contributes to the small object detection performance.
 \vspace{-0.3cm} 
\subsection{Shared Array-aware Positional Encoding (ArrayCode)}\label{sec:arraycode}
The AMIS hardware induces band-wise parallax that varies with object depth. 
Standard calibration is insufficient because depth variations in urban scenes break global homography models, and small targets amplify residual parallax. 
Instead of heavy direct feature warping, ArrayCode explicitly injects the sensor array topology into the input representation, enabling the network to learn parallax-aware features implicitly.

Let the RGB camera occupy position $\bm{p}_{\mathrm{rgb}}$ in the $3\times3$ array and band $b$ occupy cell $\bm{p}_b\in\{0,1,2\}^2$.
We define the relative geometric offset of band $b$ with respect to the RGB anchor as:
\begin{equation}
\Delta\bm{p}_b=(\Delta x_b,\Delta y_b)=\Normalize\!\left(\bm{p}_b-\bm{p}_{\mathrm{rgb}}\right)\in[-1,1]^2,
\end{equation}
where $\Normalize(\cdot)$ linearly rescales the grid offsets.
These discrete offsets are spatially expanded to form dense coordinate maps $\bm{P}_b^x, \bm{P}_b^y \in \mathbb{R}^{H\times W}$, for each band.

We formulate the MSI input as a 5D spatial-spectral tensor $\bm{X}_{\mathrm{msi}} \in \mathbb{R}^{B\times 1\times 7\times H\times W}$, where the spectral dimension corresponds to 7 bands.
To embed the physical layout, we concatenate the coordinate maps along the channel dimension, yielding a geometry-aware tensor:
\begin{equation}
	\tilde{\bm{X}}_{\mathrm{msi}}=
	\big[\bm{X}_{\mathrm{msi}},\ \bm{P}^{x},\ \bm{P}^{y}\big]
	\in \mathbb{R}^{B\times 3\times 7\times H\times W}.
\end{equation}
Here, the first channel encodes intensity, while the appended channels encode the relative displacement $(\Delta x,\Delta y)$ specific to each band.

For the RGB modality, we treat it as the geometric origin. Accordingly, we append zero-valued maps to align with the MSI feature space:
\begin{equation}
	\tilde{\bm{X}}_{\mathrm{rgb}}=
	\big[\bm{X}_{\mathrm{rgb}},\ \bm{0},\ \bm{0}\big]
	\in \mathbb{R}^{B\times 5\times H\times W},
\end{equation}
where $\tilde{\bm{X}}_{\mathrm{rgb}}$ has 3 channels and the two appended planes correspond to $(\Delta x,\Delta y)$.
Crucially, these zero planes define the RGB view as the reference frame. We refer to this reference view as the \textit{RGB anchor}, which enforces a unified spatial-spectral coordinate system across the two modalities.

ArrayCode gives the two modalities a shared spatial-spectral coordinate system derived from the physical camera layout. 
The effectiveness of this array-aware prior is further quantified in Sec.~\ref{AblationA}, where ArrayCode
consistently outperforms feature-level alignment baselines and
improves small-target detection precision.
 \vspace{-0.3cm} 
\subsection{Dual-stream Feature Extraction}
\label{sec:dual_stream}
The inherent heterogeneity between RGB spatial textures and MSI spectral signatures induces a notable modality gap. RGB operates on spatial gradients to capture geometric cues, whereas MSI models inter-band correlations to reveal material properties.
Consequently, employing a unified feature extractor is ill-suited for this task. 
A shared architecture often produces representations biased toward the dominant modality, leaving complementary cues under-exploited.
To mitigate this, we devise a dual-stream module to decouple the encoding pathways. 
We further construct hierarchical feature pyramids for both streams. 
Shallow layers retain fine-grained details to combat the weak responses of small targets. 
In contrast, deep layers encode abstract semantic context. 
This hierarchical structure establishes a robust foundation for the subsequent multi-scale fusion.

\textbf{RGB Stream: Spatial Detail Preservation.}
In the RGB stream, we adopt a DLA-34 backbone~\cite{CenterNet} that focuses on shape and texture, providing high-fidelity geometric details.
Given the ArrayCode-augmented input $\tilde{\bm{X}}_{\mathrm{rgb}} \in \mathbb{R}^{B\times 5\times H\times W}$, the network produces a three-level feature pyramid 
$\bm{F}_{\mathrm{rgb}}^{\ell} \in \mathbb{R}^{B\times (16\ell)\times \frac{H}{2^{\ell}}\times \frac{W}{2^{\ell}}}$ for $\ell \in \{1,2,3\}$, 
where $B$ denotes the batch size, $H\times W$ represents the input spatial resolution , and $\ell$ denotes the pyramid level.

\textbf{MSI Stream: Spectral Correlation Learning.}
In the MSI stream, we use a 3D encoder to model volumetric dependencies across bands.
The encoder employs 3D convolutions to capture inter-band correlations, followed by a spectral pooling operation to compress the band dimension, yielding 2D feature maps aligned with the RGB pyramid.
Given $\tilde{\bm{X}}_{\mathrm{msi}}$, the encoder generates
$\bm{F}_{\mathrm{msi}}^{\ell} \in \mathbb{R}^{B\times (16\ell)\times \frac{H}{2^{\ell}}\times \frac{W}{2^{\ell}}}$ for $\ell \in \{1,2\}$.
To identify the optimal spectral representation, we conduct a systematic evaluation of backbone architectures and band ordering strategies in Sec.~\ref{sec:axisB}, supporting the superiority of this 3D encoding approach over 2D alternatives.
 \vspace{-0.3cm} 
\subsection{Fine-scale Fusion with Semantic Decoupling}\label{sec:fusion}
\begin{table*}[t]
	\captionsetup{font=scriptsize, labelfont=scriptsize}
	\footnotesize
	\scriptsize
	\centering
	\vspace{-0.2cm} 
	\caption{
		Performance comparison on UAVNet-MS.
		All AP/AR values are reported in percentage (\%).
		ET/T/S denote object size tiers.
		$^{*}$ denotes models initialized from publicly available pretrained weights.
		OR/SH indicate ordered/shuffled spectral bands.
		$^{\dag}$ indicates baselines adapted from standard architectures \cite{CenterNet} as detailed in the Appendix Sec.~S3.2.
		\#Param.~(M), FPS, and FLOPs (G) denote the number of parameters (↓), frames per second (↑), and GigaFLOPs per image (↓), respectively.
		\textbf{Bold} indicates best, \underline{Underline} indicates second best.
	}
	\label{tab:performance_comparison}
	\vspace{-.3cm}
	\setlength{\tabcolsep}{2.5mm} 
	\setlength{\aboverulesep}{0pt}
	\setlength{\belowrulesep}{0pt}
	\renewcommand\arraystretch{1.1} 
	\begin{tabular}{cl cc ccc c ccc ccc}
		\toprule
		& \multirow{2}{*}{\textbf{Methods}} & \multirow{2}{*}{\textbf{AP$_{50}$}} & \multirow{2}{*}{\textbf{mAP}} 
		& \multicolumn{3}{c}{\textbf{AP by Size}} & & \multicolumn{3}{c}{\textbf{AR by Size}} & \multirow{2}{*}{\textbf{\#Param.}} & \multirow{2}{*}{\textbf{FPS}} & \multirow{2}{*}{\textbf{GFLOPs}} \\
		\cmidrule(lr){5-7} \cmidrule(lr){9-11}
		& & & & ET & T & S & & ET & T & S & & & \\
		\midrule
		
		\multirow{12}{*}{\rotatebox[origin=c]{90}{\textbf{RGB-only}}} 
		& Faster R-CNN*\cite{FasterR-CNN} & 28.0 & 19.1 
		& 0.0 & 4.1 & 44.9 & & 0.0 & 5.1 & \second{56.5} & 41.36 & \second{20.76} & 268.23 \\
		
		& Cascade R-CNN*\cite{CascadeR-CNN} & 22.7 & 20.4
		& 0.0 & 4.1 & \second{47.0} & & 0.0 & 4.5 & 55.3 & 69.16 & 18.48 & 342.73 \\
		
		& FCOS*\cite{FCOS} & 17.3 & 7.8 
		& 2.3 & 2.0 & 20.2 & & 15.2 & 20.1 & 38.4 & 32.12 & \best{22.90} & 299.52 \\
		
		& TOOD*\cite{TOOD} & 39.3 & \second{21.5} 
		& 9.4 & 10.9 & \best{47.6} & & 24.1 & \second{29.1} & \best{59.1} & 32.20 & 18.39 & 306.17 \\
		
		& CenterNet\cite{CenterNet} & 31.3 & 12.7 
		& 12.2 & 9.2 & 19.0 & & 31.2 & 18.1 & 30.5 & 11.70 & 6.54 & 353.60 \\
		
		& CenterNet*\cite{CenterNet} & \second{40.9} & 16.1 
		& 13.8 & \second{14.0} & 31.5 & & 36.4 & 22.2 & 41.2 & 11.70 & 6.54 & 353.60 \\
		
		& YOLOv8s\cite{YOLOv8} & 24.7 & 13.1 
		& 10.9 & 9.3 & 27.4 & & 14.2 & 14.1 & 40.1 & \second{11.40} & 10.93 & \best{72.53} \\
		
		& SAHI+YOLOv8s\cite{SAHI} & 15.7 & 8.0 
		& 3.6 & 7.2 & 25.7 & & 14.8 & 18.1 & 44.1 & \best{11.14} & 3.45 & \second{89.56} \\
		
		& YOLOv9-E\cite{YOLOv9} & 18.2 & 8.6 & 11.0 & 8.6 & 15.3 & & 16.6 & 15.4 & 27.2 & 58.15 & 6.97 & 487.74 \\
		
		& Deformable DETR*\cite{DeformableDETR} & 29.2 & 10.8 
		& 2.5 & 3.4 & 20.3 & & 10.1 & 13.7 & 36.3 & 40.01 & 14.40 & 257.80 \\
		
		& DINO\cite{DINO} & 24.7 & 9.2 
		& 13.1 & 7.6 & 12.9 & & 27.2 & 20.0 & 43.5 & 47.55 & 10.60 & 406.23 \\
		
		& Anchor DETR\cite{AnchorDETR} & 24.7 & 7.7 
		& 8.5 & 4.7 & 20.2 & & 27.5 & 18.5 & 41.8 & 30.70 & 12.66 & 251.40 \\
		\midrule
		
		\multirow{6}{*}{\rotatebox[origin=c]{90}{\textbf{MSI-only}$^{\dag}$}} 
		& 2DConv-SH & 1.8 & 0.5 & 0.2 & 0.4 & 1.5 & & 3.0 & 2.1 & 5.0 & 18.55 & 5.71 & 489.02 \\
		& 2DConv-SH* & 3.5 & 1.1 & 0.5 & 0.5 & 2.8 & & 3.2 & 2.1 & 10.0 & 18.55 & 5.71 & 489.02 \\
		& 2DConv-OR & 15.8 & 4.5 & 2.3 & 3.4 & 4.1 & & 3.3 & 2.1 & 6.9 & 18.57 & 5.41 & 489.02 \\
		& 3DGDeform-OR & 8.4 & 2.7 & 4.0 & 2.5 & 0.1 & & 5.0 & 4.1 & 1.6 & 16.90 & 5.69 & 249.69 \\
		& SCAttnSel-OR & 11.3 & 5.2 & 11.9 & 4.3 & 1.3 & & 17.9 & 8.4 & 3.9 & 15.95 & 6.13 & 249.16 \\
		& 3DConv-OR & 20.2 & 6.3 & 8.7 & 5.4 & 8.0 & & 20.7 & 9.1 & 15.6 & 16.80 & 6.45 & 246.83 \\
		\midrule
		
		\multirow{5}{*}{\rotatebox[origin=c]{90}{\textbf{Multi-modal}$^{\dag}$}} 
		& EarlySum & 8.1 & 3.2 & 1.2 & 1.8 & 2.5 & & 17.8 & 11.9 & 9.2 & 18.60 & 5.41 & 489.02 \\
		& MidConcat & 37.3 & 15.4 & 12.4 & 13.5 & 16.2 & & 31.4 & 24.2 & 34.3 & 18.69 & 4.40 & 344.14 \\
		& LateSum & 33.0 & 15.4 & \second{16.5} & 12.5 & 22.8 & & 33.8 & 26.7 & 40.1 & 16.80 & 4.46 & 432.25 \\
		& EarlyConcat & 36.7 & 15.6 & 12.1 & 13.2 & 15.9 & & \second{37.3} & 27.6 & 37.5 & 18.58 & 6.82 & 437.59 \\
		
		& \textbf{MFDNet (ours)} & \best{47.1} & \best{22.0} 
		& \best{19.9} & \best{14.7} & 37.1 
		& & \best{37.7} & \best{31.3} & 45.3 
		& 22.34 & 3.49 & 565.13 \\
		\bottomrule
	\end{tabular}
	\vspace{-.4cm}
\end{table*}
Conventional multi-scale fusion propagates cross-modal interactions across all feature levels, which can unnecessarily mix fine spectral cues with high-level semantics. For small UAVs, high-resolution features preserve local evidence for small targets, while deeper features encode semantic context beneficial for suppressing false alarms. We therefore adopt semantic decoupling, injecting spectral cues only into the feature levels where they are most informative.

Specifically, we restrict cross-modal interaction to the high-resolution pyramid levels ($\ell=1,2$). 
Aligned MSI features are injected into the RGB stream via element-wise addition:
\begin{equation}
	\bm{F}_{\mathrm{fuse}}^{\ell} = \bm{F}_{\mathrm{rgb}}^{\ell'} + \bm{F}_{\mathrm{msi}}^{\ell}, \quad \ell\in\{1,2\},
\end{equation}
effectively enhancing small-target contrast in the spatial texture maps.
Crucially, we block MSI injection at the semantic scale ($\ell=3$). 
The top-level feature $\bm{F}_{\mathrm{rgb}}^{3'}$ remains purely RGB-based, preserving a clean semantic pathway unperturbed by spectral redundancy.

Finally, the material-enhanced fine features and the semantically pure coarse features are aggregated via an IDAUp module~\cite{CenterNet} to generate the final representation $\bm{F}_{\mathrm{final}}$.
This selective fusion paradigm outperforms full-depth integration (validated in Sec.~\ref{sec:axisC}) by balancing local spectral sensitivity with global semantic stability.
The detection head follows~\cite{CenterNet}, predicting heatmaps, size, and offsets directly from $\bm{F}_{\mathrm{final}}$.
	\vspace{-.4cm}
\section{Experiments}
\subsection{Experimental Setup}
\noindent\textbf{Protocols and Metrics.}
We evaluate detection performance under three protocols: RGB-only (3-channel), MSI-only (7-band), and multi-modal (RGB+MSI). All settings use a unified ground truth to ensure rigorous comparability. We follow~\cite{Tiny24-XY,11251201} standard configurations of mAP at an IoU threshold of 0.5. Furthermore, to specifically quantify capabilities on small targets, we provide size-stratified AP and AR analysis following the scale definitions in Sec.~\ref{datasetsatis}.

\noindent\textbf{Implementation Details.}
Experiments were conducted on two NVIDIA RTX 3090 GPUs using the AdamW optimizer.
For RGB baselines, we utilize official codebases with ImageNet/COCO pretrained weights and default hyperparameters to ensure reproducibility.
For MSI and multi-modal models, we adopt consistent training schedules but tailor data augmentations by excluding color-jittering operations, thereby preserving the physical spectral correlations essential for material discrimination.
	\vspace{-.4cm}
\subsection{Benchmark Results on UAVNet-MS}
We evaluate MFDNet on UAVNet-MS against 20 detectors spanning three regimes (Table~\ref{tab:performance_comparison}): 
11 RGB-only methods (two-stage~\cite{FasterR-CNN,CascadeR-CNN}, one-stage~\cite{FCOS,TOOD,CenterNet}, YOLO-series~\cite{YOLOv8,SAHI,YOLOv9}, and Transformer-based~\cite{DeformableDETR,DINO,AnchorDETR}),
as well as MSI-only and RGB--MSI fusion baselines implemented as in-house variants (marked with $^{\dag}$), whose configurations are provided in Appendix Sec.~S3.2.

\begin{figure}[t]
	\captionsetup {font=scriptsize, labelfont=scriptsize}
	\centering
	\vspace{-0.2cm} 
	\includegraphics[width=0.46\textwidth]{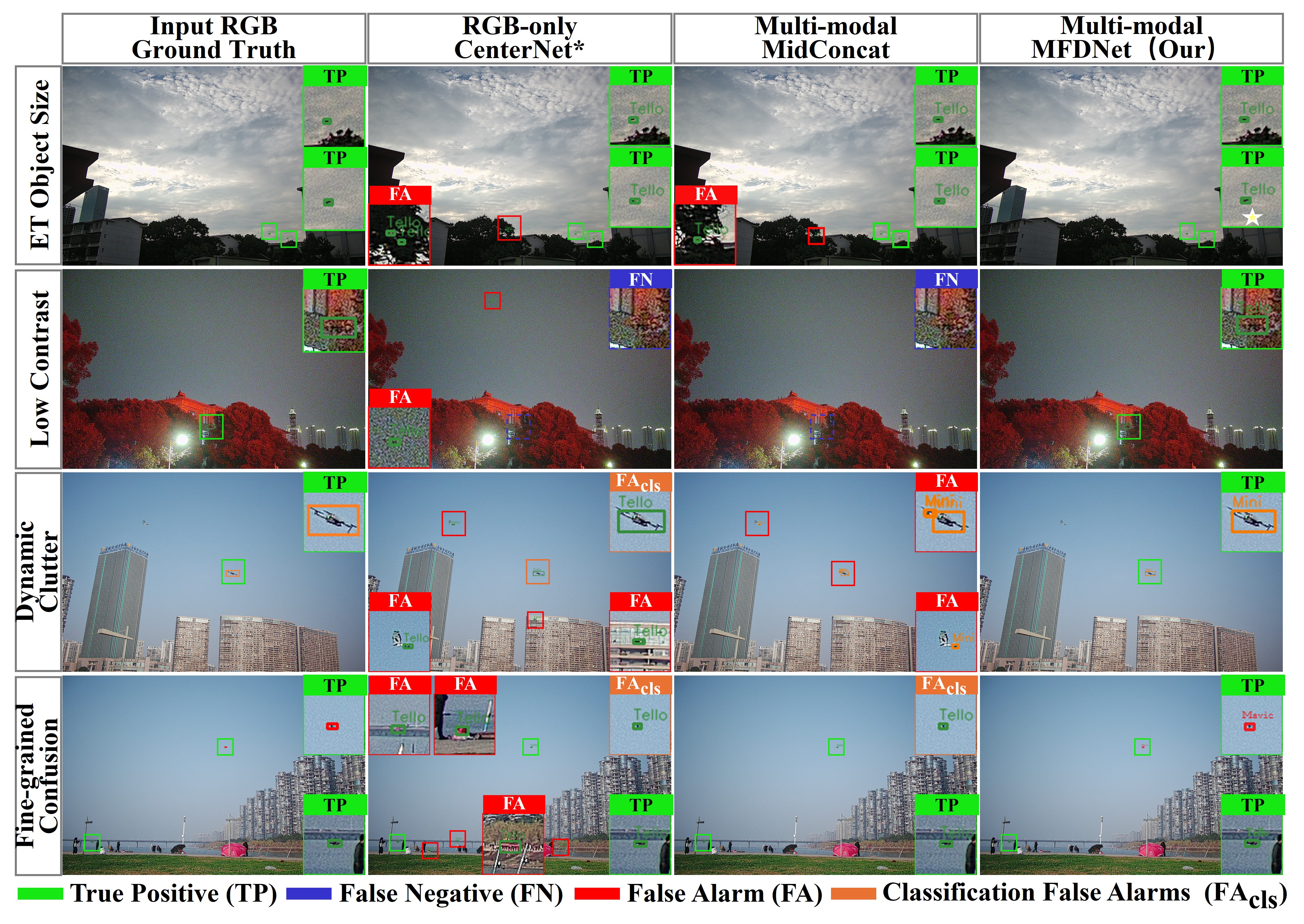}
	\vspace{-0.4cm} 
	\caption{
		Qualitative comparison under key challenges: extremely tiny objects, dynamic clutter, low contrast, and fine-grained category confusion. From left to right, columns show RGB with ground-truth annotations, CenterNet* (RGB-only), MidConcat (RGB+MSI), and MFDNet (ours).
	}
	\label{fig_compare} 
	\vspace{-.6cm}
\end{figure}
\noindent\textbf{Overall Quantitative Performance.}
Our proposed method brings a significant improvement of +6.2\% AP$_{50}$ compared to the best RGB-only detector (CenterNet*\cite{CenterNet}). 
Gains are most pronounced in the ET tier (19.9\% vs.\ 13.8\% in AP), demonstrating that spectral cues are particularly helpful when spatial features collapse.

\noindent\textbf{Qualitative Analysis under Core Challenges.}
Fig.~\ref{fig_compare} shows qualitative results under core challenges. MFDNet reduces false alarms and missed detections compared to CenterNet* and fusion methods, especially for extremely tiny or low-contrast UAVs.

\noindent\textbf{Efficiency and Architectural Justification.}
The above results position MFDNet as an accuracy-oriented baseline for the small object regime. 
This accuracy comes with a moderate increase in model size (22.34M vs. 11.70M for CenterNet*) and a lower inference speed (3.49 FPS, 565 GFLOPs per cube), reflecting the computational demand of processing high-resolution spectral volumes. 
Notably, RGB-only detectors with comparable or even larger capacity (e.g., YOLOv9-E with 58.2M model size) still fail to recover extremely tiny UAVs, whereas MFDNet attains 19.9\% AP$_{\mathrm{ET}}$ on the ET tier. 
These findings indicate that explicitly modeling spectral cues via a dual-stream design is a more effective way to improve small UAV detection than simply scaling up RGB-only architectures, and provide a solid starting point for future work on more efficient spectral fusion.
	\vspace{-0.4cm} 
\subsection{Ablation Studies}
\label{Ablation}
We conduct ablation studies on three key design choices that correspond to our architectural principles: array-aware alignment, spectral-branch design, and the fusion stage/mechanism.
	\vspace{-0.2cm}
\subsubsection{Array-aware alignment}
\label{AblationA}
\begin{figure}[t]
	\centering
	\vspace{-0.2cm} 
	\subfloat[\scriptsize Multi-modal Alignment Strategies\label{axisAa}]{%
		\includegraphics[width=0.49\linewidth]{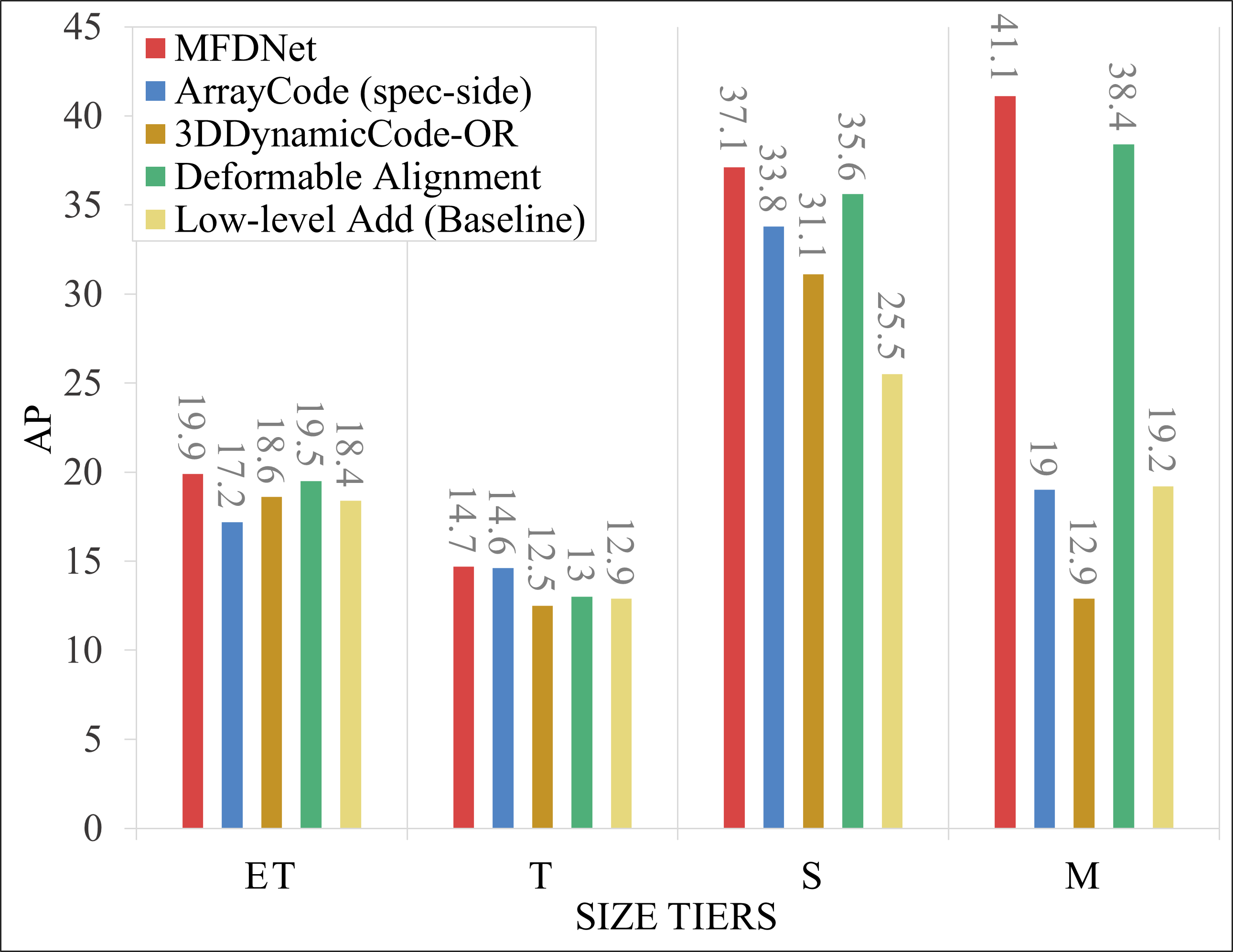}%
	}
	\hfill 
	\subfloat[\scriptsize MSI-only Alignment Strategies\label{axisAb}]{%
		\includegraphics[width=0.49\linewidth]{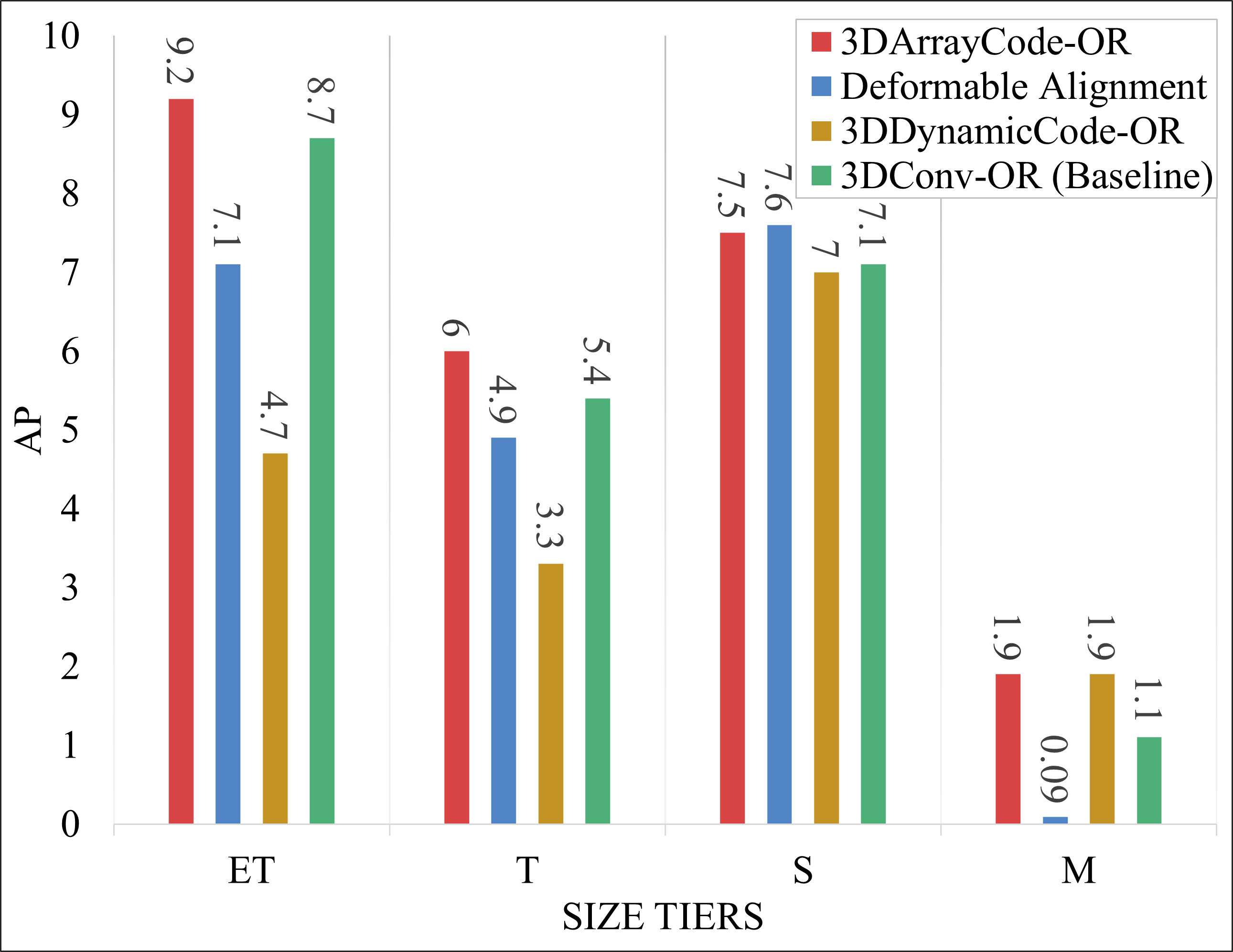}%
	}%
	\vspace{-0.2cm}
	\caption{Ablation study on alignment strategies.}
	\label{axisA}
	\vspace{-0.3cm}
\end{figure}

To validate the geometric alignment principle, we compare ArrayCode with alternative alignment modules under both multi-modal and MSI-only settings.
As shown in Fig.~\ref{axisA}, we consider three representative designs:
Deformable Alignment (feature-level deformable convolution),
3DDynamicCode (dynamic 3D warping in feature space),
and our ArrayCode family, which injects array offsets either only on the spectral side or into both RGB and MSI branches.

\textbf{Physics-driven priors vs.\ data-driven warping.}
In both multi-modal and MSI-only settings, replacing computationally intensive feature warping with ArrayCode consistently improves tiny-target accuracy. 
Specifically, compared to data‑driven warping strategies such as Deformable Alignment and 3DDynamicCode, ArrayCode‑OR consistently delivers higher or competitive AP across all size tiers, with the most pronounced gains observed on ET and T targets. This demonstrates that the physics‑driven array prior provides a more stable and effective alignment solution for small‑UAV detection.

\textbf{Importance of a shared RGB anchor.}
The symmetric variants of ArrayCode, which treats the RGB view as an anchor and MSI as offset bands, consistently outperforms the spectral-only variant.
Specifically, removing the RGB anchor reduces AP$_{\mathrm{ET}}$ by 2.7\%.
This supports that explicitly defining a shared reference frame is crucial for robust alignment.
We further show in Table~\ref{tab:axisC_fusion_stage} that these gains persist across different fusion stages, highlighting ArrayCode as a generally application, architecture-agnostic alignment prior.
\begin{table}[t]
\centering
\captionsetup{font=scriptsize, labelfont=scriptsize}
\scriptsize
\caption{Ablation of spectral-branch design choices under the MSI-only setting.}
\label{tab_axisB}
\vspace{-.4cm}
\setlength{\tabcolsep}{1.5mm}
\renewcommand\arraystretch{.98}
\begin{tabular}{lcccccccc}
\toprule
\textbf{Method} & \textbf{mAP}  &\textbf{AP$_{50}$} & \textbf{AP$_{75}$}&
\textbf{AP$_{\mathrm{ET}}$} & \textbf{AP$_{\mathrm{T}}$} & \textbf{AP$_{\mathrm{S}}$} & \textbf{AP$_{\mathrm{M}}$} \\
\midrule
2DConv-SH       & 0.5 & 1.8 &0.1& 0.2 & 0.4 & 1.5 & 2.9 \\
3DGDeform-OR    & 2.7 & 8.4 &0.7& 4.0 & 2.5 & 0.1 & 1.5 \\
2DConv-OR       & 4.5 & 15.8 &1.1& 3.9 & 3.3 & 6.9 & 7.3 \\
3DGAT-OR        & 4.7 & 14.4 & 1.1&6.0 & 3.7 & 4.1 & 0.5 \\
BandSelect-OR   & 5.1 & 15.5 &1.5& 6.0 & 4.2 & 5.2 & 1.1 \\
ArrayCode-OR    & \textbf{7.1}  &\textbf{23.9} &\textbf{1.6}& \textbf{9.2} & \textbf{6.0} & \textbf{7.5} & \textbf{7.4} \\
\bottomrule
\vspace{-0.7cm}
\end{tabular}
\end{table}
\subsubsection{Spectral branch design}
\label{sec:axisB}
We next evaluate alternative MSI spectral-branch designs under the MSI-only protocol, including band ordering and backbone variants strategies.
Table~\ref{tab_axisB} compares variants that differ in
(i) band ordering (ordered vs.\ shuffle),
(ii) encoder dimensionality (2D vs.\ 3D), and
(iii) additional feature modules (deformable convolutions, attention, band selection, and our ArrayCode-based design).
Performance spans a wide range (1.8\%-23.9\% AP$_{50}$), showing that small object spectral detection is highly sensitive to branch design.

\textbf{Preserving physical band order.}
Methods that respect the physical wavelength ordering (OR) consistently outperform their shuffled counterparts (SH).
For instance, 2DConv-OR achieves 4.5\% mAP, about $9\times$ higher than 2DConv-SH.
This supports that local spectral continuity carries meaningful structure, and MSI should not be treated as an arbitrary channel permutation of RGB.

\textbf{Benefits of 3D encoding.}
Our best 3D encoder, ArrayCode‑OR, jointly models spectral‑spatial neighborhoods and outperforms pure 2D variants (23.9\% vs. 15.8\% AP$_{50}$ for 2DConv‑OR). This is achieved by integrating ordered 3D convolutions with array‑aware priors. We attribute this to:
(i) 3D kernels following spectral response curves along the band axis, enhancing material discrimination; and
(ii) simultaneous aggregation of local spatial context—crucial when each UAV occupies only a few pixels.

\textbf{Justification for dual-stream design.}
Even with the optimal ArrayCode-OR configuration, MSI-only performance lags behind RGB-only methods. We attribute this to two structural factors intrinsic to the current spectral vision landscape:
(i) The Pre-training Gap: Unlike RGB-only detectors initialized on massive corpora, spectral backbones lack generic feature priors.
(ii) Physics of Narrow-band Imaging: Narrow spectral bands inherently capture fewer photons than broad-band RGB, resulting in lower signal-to-noise ratio (SNR) and sparse spatial texture.
These constraints support that MSI is ill-suited for independent spatial localization but excels at material differentiation. This validates our dual-stream strategy: leveraging RGB for high-fidelity spatial geometry while utilizing MSI specifically to inject fine-scale spectral correlations.
\begin{table}[t]
\centering
\vspace{-0.2cm} 
\captionsetup{font=scriptsize, labelfont=scriptsize}
\scriptsize
\caption{Analysis of fusion stages and ArrayCode (AC). "Fuse@$\ell$" injects MSI features at the $\ell$-th level ($\ell=1,2$ are fine scales, and $\ell=3$ is the semantic level). MFDNet adopts low-level fusion ($\ell=1,2$) with AC.}
\label{tab:axisC_fusion_stage}
\vspace{-.3cm}
\setlength{\tabcolsep}{0.4mm} 
\renewcommand\arraystretch{0.98} 
\begin{tabular}{l c ccc cccccc}
\toprule
\multirow{2}{*}{\textbf{Method}} & \multirow{2}{*}{\textbf{AC}} & \multicolumn{3}{c}{\textbf{Fusion Level}} & \multicolumn{2}{c}{\textbf{Accuracy}} & \multicolumn{2}{c}{\textbf{Tiny Scale}} & \multicolumn{2}{c}{\textbf{Efficiency}} \\
\cmidrule(lr){3-5} \cmidrule(lr){6-7} \cmidrule(lr){8-9} \cmidrule(lr){10-11}
& & \textbf{@$\bm{\ell}$1} & \textbf{@$\bm{\ell}$2} & \textbf{@$\bm{\ell}$3} & \textbf{mAP} & \textbf{AP$_{50}$} & \textbf{AP$_{\mathrm{ET}}$} & \textbf{AP$_{\mathrm{T}}$} & \textbf{Params} & \textbf{FPS} \\
\midrule
RGB-only & \xmark & \xmark & \xmark & \xmark & 12.7 & 31.3 & 15.2 & 9.2 & 11.7 & 6.5 \\
\midrule
\multirow{2}{*}{Partial Fusion} & \xmark & \xmark & \xmark & \cmark & 16.8 & 37.5 & 17.2 & 12.8 & 16.3 & 3.5 \\
 & \cmark & \xmark & \xmark & \cmark & 17.5 & 40.4 & \textbf{22.3} & 14.5 & 16.3 & 3.5 \\
\midrule
\multirow{2}{*}{All-level Add} & \xmark & \cmark & \cmark & \cmark & 18.0 & 32.3 & 18.0 & 12.5 & 20.2 & 3.7 \\
 & \cmark & \cmark & \cmark & \cmark & 19.2 & 43.1 & 18.2 & 14.5 &22.3& 3.4 \\
\midrule
Low-level Add & \xmark & \cmark & \cmark & \xmark & 14.6 & 34.0 & 18.4 & 12.9 & 22.3 & 3.5 \\
\textbf{MFDNet (Ours)} & \cmark & \cmark & \cmark & \xmark & \textbf{22.0} & \textbf{47.1} & 19.9 & \textbf{14.7} & 22.3 & 3.5 \\
\bottomrule
\end{tabular}
\vspace{-0.4cm}
\end{table}
\begin{table}[t]
\centering
\captionsetup{font=scriptsize, labelfont=scriptsize}
\scriptsize
\caption{Comparison of different fusion mechanisms at the fine-scale stage.}
\label{tab:axisC_fusion_mechanism}
\vspace{-.3cm}
\setlength{\tabcolsep}{1.8mm}
\renewcommand\arraystretch{.98}
\begin{tabular}{lccccccc}
\toprule
\textbf{Method} & \textbf{mAP} & \textbf{AP$_{50}$} & \textbf{AP$_{75}$} & \textbf{AP$_{\mathrm{ET}}$} & \textbf{AP$_{\mathrm{T}}$} & \textbf{AP$_{\mathrm{S}}$} & \textbf{AP$_{\mathrm{M}}$} \\
\midrule
Low-level Add     & 14.6 & 34.0 & 9.5 & 18.4 & 12.9 & 25.5 & 19.2 \\
MidAttn           & 17.8 & 40.9 & 10.3 & 19.6 & 13.5 & 28.3 & 20.6 \\
MidCrossAttn         & 17.8 & 38.6 & 12.3 & 18.0 & 14.6 & 35.7 & 19.2 \\
MidDeforAttn   & 16.3 & 37.6 & 9.2 & 18.6 & 12.5 & 31.1 & 12.9 \\
MidGateAttn     & 16.6 & 37.0 & 11.4 & 17.4 & 11.6 & 32.7 & 23.3 \\
MFDNet               & \textbf{22.0} & \textbf{47.1} & \textbf{16.2} & \textbf{19.9} & \textbf{14.7} & \textbf{37.1} & \textbf{41.1} \\
\bottomrule
\vspace{-0.8cm}
\end{tabular}
\end{table}
\vspace{-0.1cm}
\subsubsection{Fusion stage and mechanism}
\label{sec:axisC}
Finally, we examine how fusion design affects RGB--MSI small object detection, in terms of both where to fuse (stage) and how to fuse (mechanism).

\textbf{Impact of fusion level.}
Table~\ref{tab:axisC_fusion_stage} compares four fusion layouts. 
Fusing MSI at the top semantic layer ($\ell=3$) tends to hurt overall accuracy (mAP/AP$_{50}$) and tiny-tier performance, whereas restricting fusion to fine scales ($\ell=1,2$) as in MFDNet yields the best trade-off.

\textbf{Impact of fusion mechanism.}
At the same fusion stage (Low-level Add), Table~\ref{tab:axisC_fusion_mechanism} compares additive fusion with heavier attention-based variants. Adding deformable alignment, cross-attention, or gating on low-level fusion yields no consistent gains and often degrades AP$_{75}$ due to feature distortion from repeated resampling. MFDNet’s minimalist design instead achieves the best results across all metrics, showing that lightweight, stable spectral injection is preferable to complex cross-attention in this low‑SNR, small‑object regime.
\vspace{-0.5cm}
\subsection{Robustness Across Conditions}
\label{sec:condition}
\begin{figure}[t]
	\centering
		\vspace{-0.5cm}
	\subfloat[\scriptsize mAP Comparison\label{condition_a}]{%
		\includegraphics[width=0.5\linewidth]{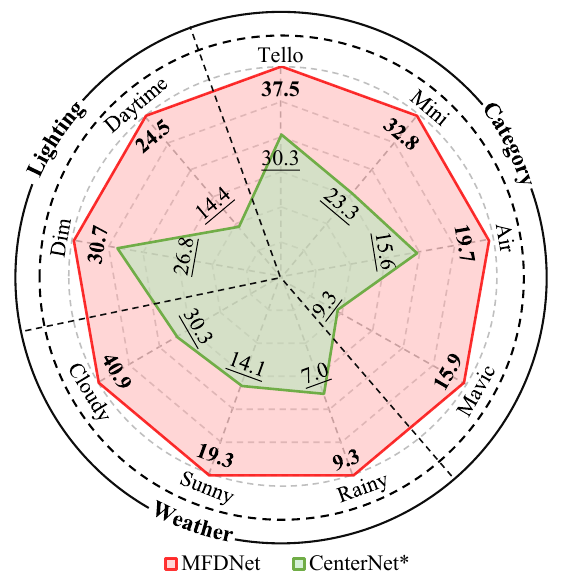}%
	}%
	\hfill 
	\subfloat[\scriptsize mAR Comparison\label{condition_b}]{%
		\includegraphics[width=0.5\linewidth]{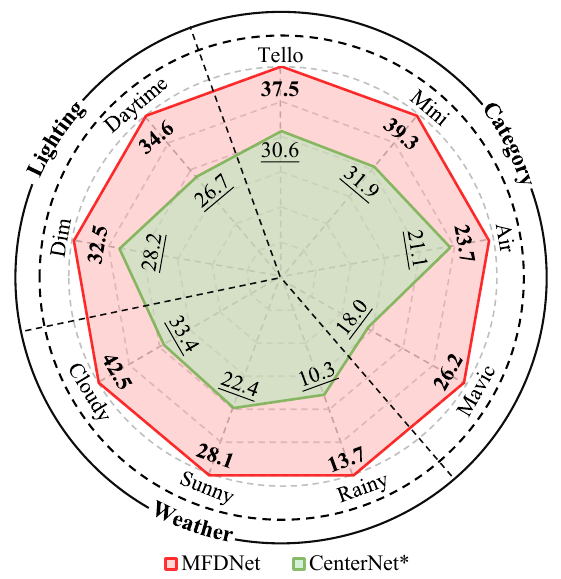}%
	}%
	\vspace{-0.2cm}
	\caption{Robustness of MFDNet across conditions.}
	\label{condition}
	\vspace{-0.6cm}
\end{figure}

We assess robustness across conditions via radar plots (Fig.~\ref{condition}). Subplots (a) and (b) report mAP and mAR, respectively, for both CenterNet (RGB-only) and MFDNet. We select CenterNet as a strong RGB-only baseline due to its backbone compatibility with our RGB stream. MFDNet consistently forms the outer envelope across all axes in both metrics, indicating balanced gains in accuracy and recall. MSI cues provide complementary information beyond RGB, leading to performance improvements.
\vspace{-0.5cm}
\section{Conclusion}
In this paper, we advance fine-grained small-UAV detection for low-altitude UAV monitoring by introducing a multispectral benchmark and baseline.
We introduce UAVNet-MS, a material-aware multispectral benchmark with temporally synchronized, high-resolution RGB--MSI observations to support fine-grained small-UAV detection.
To bridge array-based multispectral sensing and fine-grained small-UAV detection, we propose MFDNet as a dual-stream baseline.
With ArrayCode and scale-aware semantic-decoupled fusion, MFDNet injects fine-scale spectral cues to enhance small-target contrast while preserving stable high-level RGB semantics for false-alarm suppression.
Extensive experiments show that MFDNet consistently outperforms RGB-only baselines on UAVNet-MS, especially in the ET tier, demonstrating practical value of multispectral material cues.

Beyond the reported gains, our analyses highlight two future directions.
First, the remaining gap between RGB-only and spectral baselines motivates stronger spectral-oriented representation learning and fusion strategies, which UAVNet-MS can benchmark systematically.
Second, while volumetric spectral modeling is effective, its computational cost calls for lightweight spectral computing for resource-constrained deployment.
We hope UAVNet-MS and MFDNet provide a strong benchmark and baseline to facilitate future MSI-based UAV monitoring research.

\scriptsize
\bibliographystyle{IEEEtran}
\bibliography{egbib.bib}
\end{document}